\documentclass[10pt,twocolumn,letterpaper]{article}

\usepackage{times}
\usepackage{epsfig}
\usepackage{graphicx}
\usepackage{amsmath}
\usepackage{url}
\usepackage{amssymb}
\usepackage[ruled,linesnumbered]{algorithm2e}
\usepackage{float}
\usepackage{comment}
\usepackage{fixltx2e}
 \usepackage{multirow}
\usepackage{caption}
\newcommand{\R}{\mathbb{R}}
\newcommand{\C}{\mathbb{C}}

\begin{document}

\title{Progressive Gradient Pruning for Classification, Detection and Domain Adaptation}



\author{Le Thanh Nguyen-Meidine, Eric Granger, Marco Pedersoli, Madhu Kiran\\
LIVIA, Dept. of Systems Engineering\\
Ecole de Technologie Superieur\\
{\tt\small le-thanh.nguyen-meidine.1@ens.etsmtl.ca}
\and
Louis-Antoine\\
Genetec Inc.\\
Montreal, Canada\\
{\tt\small lablaismorin@genetec.com}
}
\maketitle

\begin{abstract}
   Although deep neural networks (NNs) have achieved state-of-the-art accuracy in many visual recognition tasks, the growing computational complexity and energy consumption of networks remains an issue, especially for applications on platforms with limited resources and requiring real-time processing. Filter pruning techniques have recently shown promising results for the compression and acceleration of convolutional NNs (CNNs). 
   However, these techniques involve numerous steps and complex optimisations because some only prune after training CNNs, while others prune from scratch during training by integrating sparsity constraints or modifying the loss function. 

   In this paper we propose a new Progressive Gradient Pruning (PGP) technique for iterative filter pruning during training. 
   In contrast to previous progressive pruning techniques,
   it relies on a novel filter selection criterion that measures the change in filter weights, uses a new hard and soft pruning strategy and  effectively adapts momentum tensors during the backward propagation pass.  
   Experimental results obtained after training various CNNs on image data for classification, object detection and domain adaptation benchmarks indicate that the PGP technique can achieve a better trade-off between classification accuracy and network (time and memory) complexity than PSFP and other state-of-the-art filter pruning techniques.
\end{abstract}

\vspace{-5mm}
\section{Introduction}

Convolutional neural networks (CNNs) learn discriminant feature representations from labeled training data, and have achieved state-of-the-art accuracy across a wide range of visual recognition tasks, e.g., image classification, object detection, and assisted medical diagnosis. Since the breakthrough results achieved with AlexNet for the 2012 ImageNet Challenge \cite{Alexnet}, CNN's accuracy has been continually improved with architectures like VGG \cite{VGG}, ResNet \cite{Resnet} and DenseNet \cite{Densenet}, at the expense of growing complexity (deeper and wider networks)  that require more training samples and computational resources \cite{SpeedAndAcc}. In particular, the speed of the CNNs can significantly degrade with such increased complexity. 

In order to deploy these powerful CNN architectures on compact platforms with limited resources (e.g., embedded systems, mobile phones, portable devices) and for real-time processing (e.g., video surveillance and monitoring, virtual reality), the time and memory complexity and energy consumption of CNNs should be reduced. This is particularly true when the model also has to adapt to a new environment/domain. For instance, the application of CNN-based architectures to real-time face detection in video surveillance remains a challenging task \cite{Ipta2017}, because of the complexity of real-time detection and the handling of different scenes, environments and camera angles. Currently, the more accurate detectors such as region proposal networks are too slow for real-time applications \cite{FRCNN, RFCN},  faster detectors such as single-shot detectors are less accurate \cite{SSD, Yolo} and none of these are universal detector that can work on any new environment without domain adaptation.
Consequently, effective methods to accelerate and compress deep networks, especially during training/domain adaptation are required to provide a reasonable trade-off between accuracy, efficiency and deployment time. 

This paper focuses on filter-level pruning techniques. While it does not provide the compression level of unstructured pruning, the reduction of parameters can be converted in a real speed up  while preserving network accuracy \cite{L1Pruning,Molchanov}. These techniques attempt to remove the filters and input channels at each convolution layer using various criteria based on, e.g.,  L1 norm \cite{L1Pruning}, or the product of feature maps and gradients\cite{Molchanov}. 
Pruning techniques can be applied under two different scenarios: either (1) from a pre-trained network, or (2) from scratch. In the first scenario, pruning is applied as a post-processing procedure, once the network has already been trained, through an one-time pruning (followed by fine-tuning) \cite{L1Pruning} or complex iterative process \cite{Molchanov} using a validation dataset \cite{L1Pruning, Entropy}, or by minimizing the reconstruction error \cite{ThiNet}. In the second scenario, pruning is applied from scratch by introducing sparsity constraints and/or modifying the loss function to train the network \cite{Network_Slimming, Learning, Discrimination-aware}. The later scenario can have more difficulty converging to accurate network solutions (due to the modified loss function), and thereby increase the computational complexity required for the optimisation process. For greater training efficiency, the progressive soft filter pruning (PSFP) method was recently introduced \cite{PSP}, allowing for iterative pruning from scratch, where filters are set to zero (instead of removing them) so that the network can preserve a greater learning capacity.  This method, however, does not account for the optimization of soft pruned weights which can have an negative impact on accuracy, because pruned weights are still being optimized with old momentum values accumulated from previous epochs.

 
In this paper, a new Progressive Gradient-based Pruning (PGP) technique is proposed for iterative filter pruning to provide a better trade-off between accuracy and complexity. To this end, the filters are efficiently pruned in a progressive fashion while training a network from scratch, and accuracy is maintained without requiring validation data and additional optimisation constraints. In particular, PGP improves on PSFP by integrating hard and soft pruning strategies to effectively adapt the momentum tensor during the backward propagation pass. It also integrates an improved version of the Taylor selection criterion \cite{Molchanov} that relies on the gradient w.r.t weights (instead of output feature maps), and is more suitable for progressive filter-based pruning. For performance evaluation, the accuracy and complexity of proposed and state-of-the-art filter pruning techniques are compared using Resnet, LeNet and VGG networks trained to address benchmark image classification (MNIST and CIFAR10 datasets), object detection (PASCAL VOC dataset) and domain adaptation (Office-31) problems. From our experiments we show that the proposed approach performs comparably or better than most of the previous techniques on image classification as well as object detection. 
Additionally, we also found that by performing pruning and domain adaptation jointly our method outperforms other approaches based on two separate steps.

\section{Compression and Acceleration of CNNs}

In general, time complexity of a CNN depends more on the convolutional layers, while the fully connected layers contain the most of the number of parameters. 
Therefore, the CNN acceleration methods typically target lowering the complexity of the convolutional layers, while the compression methods usually target reduced complexity of the fully connected layers\cite{Huffman, First_pruning_W}. This section provides an overview of the recent acceleration and compression approaches for CNNs, namely,  quantization,  low-rank approximation, 
and network pruning. Finally, a brief survey on the filter pruning methods and challenges is presented.

\subsection{Overview of methods:}

\paragraph{Quantization:} A deep neural network can be accelerated by reducing the precision of its parameters. Such techniques are often used on general embedded systems, where low-precision, e.g., 8-bit integer, provides  faster processing than the higher-precision representation, e.g.,  32-bit floating point. There are two main approaches to quantizing a neural network -- the first focuses on quantizing using weights\cite{Huffman, INQ}, and the second uses both weights and activations for quantization \cite{SYQ, BinaryNet}. These techniques can be either scalable \cite{Huffman, INQ} or non-scalable \cite{HWGQ, SYQ, BinaryNet, XNOR}, where scalable techniques means that an already quantized network can be further compressed. 

\vspace{-0.4cm}
\paragraph{Low-rank decomposition:} Low-rank approximation (LRA) can accelerate CNNs by decomposing a tensor in lower rank approximations as vector products.  \cite{LRA_Jader, LRA, LRD_FTCP}.
There are different ways of decomposing convolution tensor. Techniques like \cite{LRA_Jader, LRA} focus on approximating tensor by low rank tesnor that can be  obtained either in a layer by layer fashion \cite{LRA_Jader} or by scanning the whole network \cite{LRA}. \cite{Coordinating} proposes to force filers to coordinate more information into a lower rank space during training and then decompose it once the model is trained. Another technique employed the CP-Decomposition (Canonical Polyadic Decomposition), where a good trade-off between accuracy and efficiency is achieved \cite{LRD_FTCP}.

\vspace{-0.4cm}
\paragraph{Pruning:} Pruning is a family of techniques that removes non-useful parameters from a neural network. There are several approaches of pruning for deep neural networks. The first is weight pruning, where individual weights are pruned. This approach  has proven to  significantly compress and accelerates deep neural networks  \cite{Huffman, PruningW_ADDM, First_pruning_W}. Weight pruning techniques  usually employ sparse convolution algorithms \cite{SparseCNN, SBNet}.
The other approach is output channel or filter pruning, where  complete output channel or filters are pruned \cite{L1Pruning, ThiNet, PSP, Discrimination-aware}.  Since this paper proposes a method for filter pruning, we provide more details on this approach in the next section.

\subsection{Filter pruning:}

Following the work of Optical Brain Damage \cite{LecunnPrune}, one of the first papers that showed the efficiency of filter-level pruning  was \cite{L1Pruning}, where the weight norm is used to identify and prune weak filters, filters that do not contribute much to network. 
Afterwards,  several works proposed pruning procedures and filter importance metrics. These methods can be organized in five  pruning approaches: 1) Pruning as  one time post processing and then fine tune-- this approach is simple and easy to apply  \cite{L1Pruning}, 2) Pruning in an iterative way once the model was trained-- the iterative pruning and fine-tune increase the chance of recovering accuracy loss directly after a filter is pruned \cite{Molchanov, Molchanov_2019_CVPR}, 3) Pruning by minimizing the reconstruction error-- minimizing the reconstruction error at each layer allows the model to approximate the original performance \cite{ThiNet, Lasso, Discrimination-aware}, 
4) Pruning by using sparse constraints with a modified objective function--  to let the network consider pruning during optimization \cite{Network_Slimming, Compression-aware_Training_of_Deep_Networks, Learning_the_Number_of_Neurons_in_Deep_Networks, Lemaire_2019_CVPR}, 5) Pruning progressively while training from scratch or pre-trained model --  soft pruning \cite{SFP, He_2019_CVPR} was applied where filters are set to zero instead of actually removing them (hard manner), which leaves the network with more capacity to learn \cite{PSP}.

While first three approaches are capable of reducing the complexity of a model, they are only applied when the model is already trained, it would certainly be more beneficial to be able to start pruning from scratch during training. While, the fourth approach can start the pruning from scratch by adding sparse constraints and by modifying the optimization objective, this makes the loss harder and more sensitive to optimize. 
This can be potentially complicated when the original loss function is hard to optimize since this type of approach modifies the original loss function therefore making it potentially harder for the model to converge to a good solution.  
The fifth approach eases this process by not removing filters and uses the original loss function. However, we think that this approach can be improved since, currently, this approach does not handle pruning in the backward pass and only set the weak filters to zero. Also, the current approach calculates the L2 criterion separately from when the parameters are updated, i.e. not when we are iterating inside an epoch. For our approach we want to directly compute the criterion during update, i.e. when we are iterating in an epoch and updating parameters. 

Another important part of pruning filters is the capacity to evaluate the importance of a filter. In literature, there has been a lot of criteria that has been used to evaluate the importance of filters, e.g. L1 \cite{L1Pruning}, APoz \cite{APoz}, Entropy \cite{Entropy}, L2 \cite{PSP} and Taylor \cite{Molchanov}. Among these, the Taylor criterion \cite{Molchanov} has the most potential for pruning during training since the criterion is the result of trying to minimize the impact of having a filter pruned.

\section{Progressive Gradient Pruning}

\subsection{Pruning strategy with momentum:}

In a regular CNN, the weight tensor of a convolutional layer $l$ can be defined as $\textbf{W} \in \R^{n_\text{out} \times n_\text{in} \times k \times k}$, where $n_\text{in}$ and $n_\text{out}$ are the number of input and output channels (filters), respectively. A weight tensor of filter $i$ can be then defined as $\textbf{W}_{i} \in \R^{n_\text{in} \times k \times k}$. In order to select the weak filters of a layer, we evaluate the importance of an filter using a criterion function $c$, is usually defined as $c(\textbf{W}_i): \R^{n_\text{in} \times k \times k} \xrightarrow{} \R$. Given an filter, it yields a scalar that represents the rank, e.g. L1 \cite{L1Pruning} or gradient norm in our case.  

In order to prune convolution layer progressively, an exponential decay function is defined such that there is always a solution in  $\R$. (It is slightly different than in \cite{PSP}, where the decay function can have solutions in $\C$.) This decay function allows to select the number of weak filters at each epoch. The decay function is defined as the ratio of filters remaining after the training on epoch $t$:
\begin{equation}\label{Exponential decay function}
	\begin{aligned}
	p_t = \exp \left( \frac{\log(1 - t_\text{prune})}{T}  t \right) , 
	\end{aligned}
\end{equation}
where $t_\text{prune}$ is a hyper-parameter that defines the ratio of filters to be pruned, and $t \in \{1, 2,\ldots,T\}$ is the epoch. Since we progressively prune layer by layer and epoch by epoch, we calculate the the number of weak filters or the number of remaining filters at each layer, $n_\text{wc}$. Given ratio $p_t$ at epoch $t$, the number of weak filters for any layer is defined as:
\begin{equation}
    \label{Number of weaks Equation}
	\begin{aligned}
	n_\text{wc} = n_{c}  ( 1 - p_t ) \ ,
	\end{aligned}
\end{equation}
where $n_{c}$ can be the original number of filters of any layers. Using the the number of weak filters $n_\text{wc}$ and a pruing criterion function $c$, we end up having a subset of filters $\textbf{W}_\text{weak} \in \R^{n_\text{wc} \times n_\text{in} \times k \times k}$ with the smallest value. This subset is further divided into two subsets, using a hyper-parameter $r$ that decides the ratio of hard-to-remove filters. The subset $\textbf{W}_\text{rh} \in \R^{(n_\text{wc} \cdot r) \times n_\text{in} \times k \times k}$ is removed completely, while  the subset $\textbf{W}_\text{rs} \in \R^{n_\text{wc} \cdot (1 - r) \times n_\text{in} \times k \times k}$ will be reset to zero while keeping $R_h$ and $R_s$ as indexes for the backward pass. Additionally, hard pruning is performed on the input channels of the next layer using $R_h$.

\begin{figure}[h!]
    \centering
    \includegraphics[width=6.5cm]{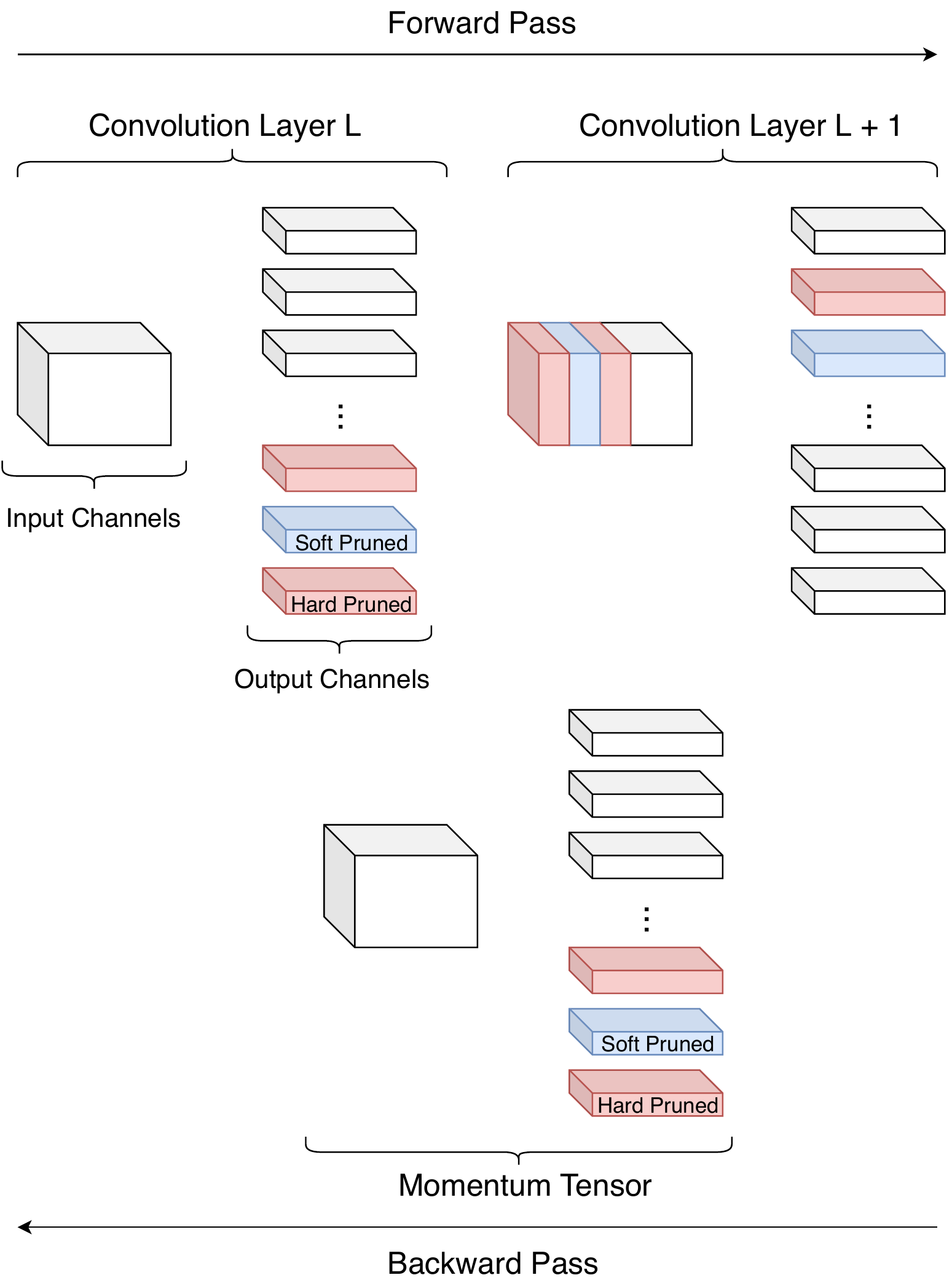}
    \caption{Illustration of the PGP pruning strategy between two successive convolutional layers.}
    \label{Pruning forward and backward}
    \vspace{-4mm}
\end{figure}

\begin{figure}[t!]
    \centering
    \includegraphics[width=6.7cm]{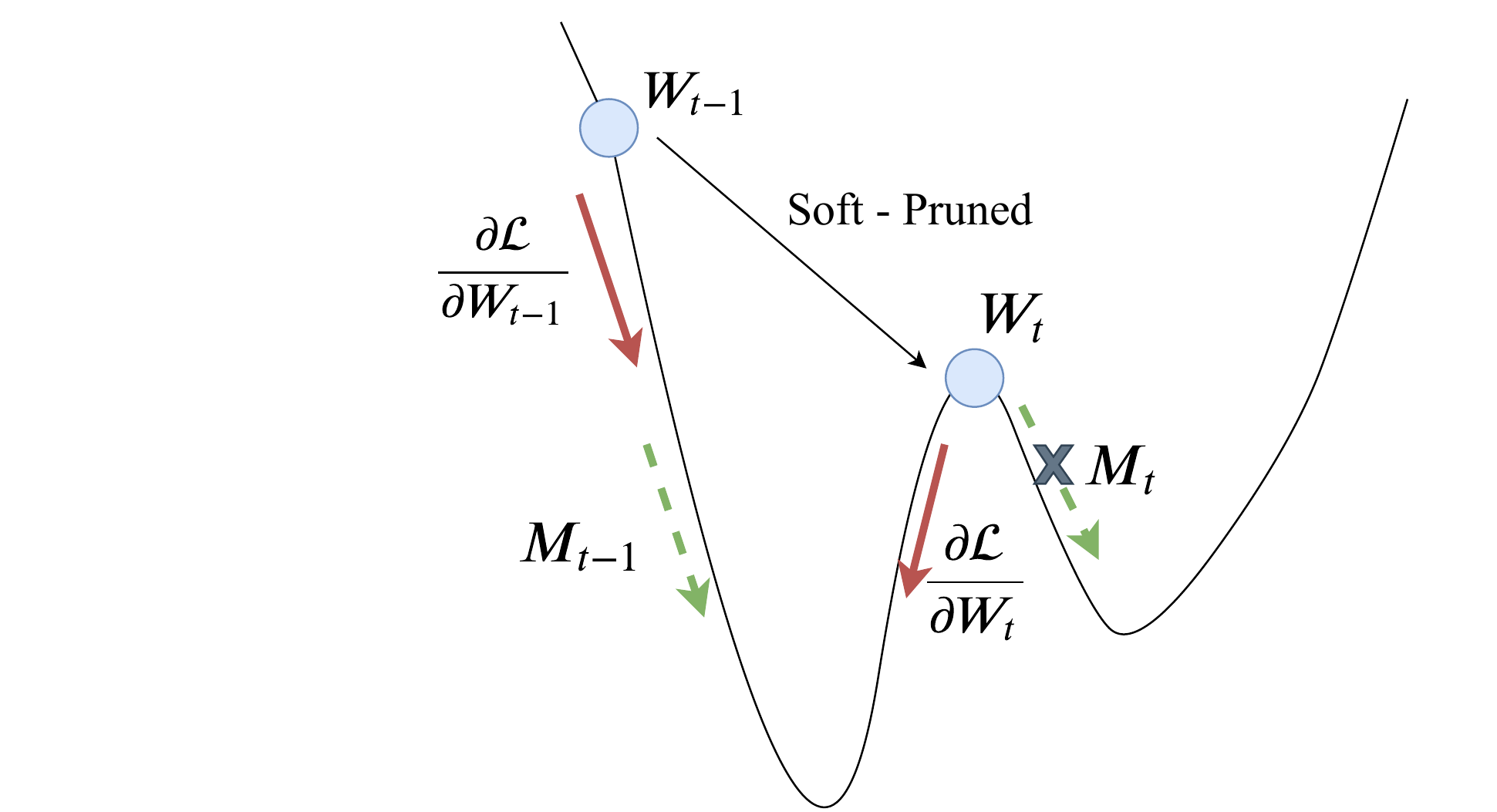}
    \caption{An illustration of the optimization process of a weight tensor $W_t$ during a progressive pruning with soft and momentum pruning. The dotted green line indicates the direction of the momentum, while the red full line indicates the direction of the gradient. At iteration $t$ the weight tensor $W_t$ is soft pruned. If the momentum tensor $M_t$ is not soft pruned, even if the gradient direction of $W_t$ is correct, the old momentum would force it to follow another direction.}
    \label{fig:Momentum pruning}
\vspace{-4mm}
\end{figure}

Figure~\ref{Pruning forward and backward} illustrates the hard and soft pruning strategy of the PGP technique, with the momentum tensor defined as $\textbf{M} \in \R^{n_\text{out} \times n_\text{in} \times k \times k}$, same dimension as a weight tensor. Using the indexes of $R_s$, we set to zero the subset $\textbf{M}_\text{rs} \in \R^{\dim(R_s) \times n_\text{in} \times k \times k}$ and hard prune the subset $\textbf{M}_\text{rh} \in \R^{\dim(R_h) \times n_\text{in} \times k \times k}$ using indexes $R_h$. Currently, progressive pruning techniques like \cite{PSP}, only the weights set to zero during training, without handling the previously-accumulated momentum accumulated which is critical for the optimization. As illustrated in Figure~\ref{fig:Momentum pruning}, momentum pruning is important for the optimization process.

Let us take a closer look at the typical equations for update of weight and momentum:
\begin{equation}\label{eq:update_weight}
	\begin{aligned}
        \textbf{W}_{t+1} = \textbf{W}_{t} - \alpha \textbf{M}_t
	\end{aligned}
\end{equation}
\begin{equation}\label{eq:momentum}
	\begin{aligned}
        \textbf{M}_t = \beta \textbf{M}_{t-1} + (1- \beta) \frac{\partial \mathcal{L}}{\partial \textbf{W}_t}
	\end{aligned}
\end{equation}
\noindent where $\textbf{W}_t$ and $\textbf{M}_t$ are respectively the weight and momentum tensors at iteration $t$, and $\alpha$ and $\beta$ are the learning rate and momentum hyper-parameters, respectively. By expanding $M_{t-1}$ in Equ. \ref{eq:momentum}:
\begin{equation}\label{eq:momentum_exp}
	\begin{aligned}
        \textbf{M}_t 
         &= \beta  (\beta \textbf{M}_{t-2} + (1 - \beta) \frac{\partial \mathcal{L}}{\partial \textbf{W}_{t- 1}}) + (1- \beta) \frac{\partial \mathcal{L}}{\partial \textbf{W}_t}
	\end{aligned}
\end{equation}
The tensor $\textbf{M}_{t-1}$ depends on the previous gradient of weight at time $t-1$. Using a soft pruning technique (like PSFP), the momentum tensor $\textbf{M}_{t-1}$ using $\frac{\partial \mathcal{L}}{\partial \textbf{W}_{t- 1}}$ is meaningless if $\textbf{W}$ is soft pruned at $t$, since the weight is reset, meaning the optimization point is no longer the same. It is therefore important to adapt the momentum tensor during soft pruning.  Our solution is to perform soft prune the momentum such that the weight tensor is correctly optimized. 

\subsection{Selection criteria:}

Molchanov at al. \cite{Molchanov} proposed the following criterion $|\Delta \mathcal{L}(\textbf{H}_i)|$  to measure the importance of a feature map $\textbf{H}_i$ from a filter $\textbf{W}_i$, computed at each layer, and for each filter:  
\begin{equation}\label{Molchanov Equation}
	\begin{aligned}
	\big|\Delta \mathcal{L}(\textbf{H}_i)\big| =  \scalebox{1}{$\big|\mathcal{L}(\mathcal{D}|\textbf{H}_i = 0) - \mathcal{L}(\mathcal{D}|\textbf{H}_i)\big|$} \approx	 \scalebox{1}{$\big|\frac{\partial \mathcal{L}}{\partial \textbf{H}_i}  \textbf{H}_i \big|$}
	\end{aligned}
\end{equation}
The term $\mathcal{L}(\mathcal{D}|\textbf{H}_i = 0)$ refers to the loss of a model when a labeled dataset $D$ is given with a pruned feature map $\textbf{H}_i = 0$. $\mathcal{L}(\mathcal{D}|\textbf{H}_i)$ is the original loss before the model has been pruned. In summary, the criterion of Equ. \ref{Molchanov Equation} is the difference between the loss of a pruned model and the original model. The criterion grows with the impact the feature map. This criterion has been shown to work well on some trained network. However, in the scenario where the network is pruned from scratch, we argue that information measured from feature map $\bf{H_i}$ is not informative since the model is not trained. Empirical results in Section 4 also support that the criterion of Equ. \ref{Molchanov Equation} is not effective at other criteria for progressive pruning.

Instead of using  $\textbf{H}_i = 0$ to prune a feature map \cite{MeanGradient} or filter, we can replace $\textbf{H}_i$ with $\textbf{W}_i$ since setting an filter to zero is the same as pruning it \cite{PSP}. The same Taylor expansion from \cite{Molchanov} then can applied with $\textbf{W}_i$, resulting in:
\begin{equation}\label{Taylor Weight}
	\begin{aligned}
	\text{TW} =  \scalebox{1}{$\big|\mathcal{L}(\mathcal{D}|\textbf{W}_i = 0) - \mathcal{L}(\mathcal{D}|\textbf{W}_i)\big|$} \approx	 \scalebox{1}{$\big|\frac{\partial \mathcal{L}}{\partial \textbf{W}_i}  \textbf{W}_i\big|$}
	\end{aligned}
\end{equation}
Equ. \ref{Taylor Weight} can be further simplified when taking in account the soft pruning nature. We can decomposed this equation because $|\frac{\partial \mathcal{L}}{\partial \textbf{W}_i}\textbf{W}_i|$ is an element-wise multiplication:
\begin{equation}\label{Taylor Weight Decomposed}
	\begin{aligned}
	\mbox{$\big|\frac{\partial \mathcal{L}}{\partial \textbf{W}_i}\textbf{W}_i\big|$} =  \scalebox{1}{$\big|\frac{\partial \mathcal{L}}{\partial \textbf{W}_i}\big|\big|\textbf{W}_i\big|$}
	\end{aligned}
\end{equation}
where $|\textbf{W}_i|$ is the absolute value of the weight of filter $i$. This meant that $|\textbf{W}_i|$ can be or very close to zero if $\textbf{W}_i$ was one of the filter that was soft-pruned. In this case, $\textbf{W}_i$ has little chance to recover, since it will likely be pruned. In order to encourage more recovery on soft prune filters, we propose to remove the $|\textbf{W}_i|$ term: 
 \begin{equation}\label{Gradient Norm}
	\begin{aligned}
	\text{GN}_i =  \scalebox{1}{$\big|\frac{\partial \mathcal{L}}{\partial \textbf{W}_i}\big|$} 
	\end{aligned}
\end{equation}
where $\text{GN}_i$ is the criterion for our approach for $i$ filter. There are two ways of calculating our criterion: 
\begin{itemize}
   \item PGP: performs a training epoch without updating the model, and compute the pruning criterion. This amounts to a batch gradient descent without updating the parameters at then end, and can provide better performance since the optimization is less noisy than SGD.
    \item RPGP: computes the pruning criterion directly during a forward-backward pass of training (while updating). This approach uses a SGD optimizer and calculates the criterion directly during the optimization and update of the model.
\end{itemize}

In either case, the criterion is applied over several iterations, so there are two ways of interpreting Equ. \ref{Gradient Norm}. One natural way of interpreting is by accumulating gradients, where the gradients are summed up to the total gradient of an filter. Since PGP goes thought the entire epoch without updates. We can use an L1 norm in order to sum up the variation inside an filter using criterion:
 \begin{equation}\label{Gradient Norm - GD}
	\begin{aligned}
	\text{$GN_G$}^i = \scalebox{1}{$\big|\big|\sum_j^N{\textbf{G}_{ij}}\big|\big|_1$}
	\end{aligned}
\end{equation}
where $\textbf{G}_{ij}$ is the gradient tensor of an filter $i$ at iteration $j$ inside an epoch. Equ. \ref{Gradient Norm - GD} measures the amount of global changes for an filter at the end of an epoch, which makes it most suitable for PGP.  The second way of interpreting is by accumulating the actual changes of an filter at each updates, using criterion:
\begin{equation}\label{Gradient Norm - SGD}
	\begin{aligned}
	\text{$GN_S$}^i =  \scalebox{1}{$\sum_j^N{||\textbf{G}_{ij}}||_1$}
	\end{aligned}
\end{equation}
Equ. \ref{Gradient Norm - SGD} calculates the L1 norm of a gradient tensor of an filter at each iteration during an epoch. Thus, instead of measuring the global change only at the end like Equation \ref{Gradient Norm - GD}, this measure the gradual changes during an epoch. This criteria is most suitable  for RPGP since the weight is updated at the same time as we accumulate our gradient.  PGP is summarized in Algo. \ref{PGP Pruning algorithm}. The algorithm for RPGP is similar but the criterion is calculated directly at the train step.

\begin{algorithm}[h]
\SetAlgoLined
\caption{Progressive Gradient Pruning method.}
\label{PGP Pruning algorithm}
\SetKwInOut{Input}{input}
\SetKwInOut{Output}{output}
\SetKwInOut{Parameter}{parameter}
\Input{A non-trained model $m$, a target percent of pruned away $t_\text{prune}$, remove ratio $r$, number of epochs $T$}
\Output{Pruned trained model}

\For {$t\gets{1}$ \KwTo {T}} {
    Train the model for one epoch\\
    \ForEach {convolution layer $C_l$} {
        Calculate the number of weak filters $n_\text{wc}$ (\ref{Number of weaks Equation})\\
        Calculate the pruning criterion using $GN_G$ (\ref{Gradient Norm - GD}) or $GN_S$ (\ref{Gradient Norm - SGD})  \\
        Partition $\textbf{W}_\text{weak}$ into indexes $R_h$ (hard remove filters) and $R_s$ (soft remove filters) using $r$\\
        Remove subset $\textbf{W}_{rh}$ and set $\textbf{W}_\text{rs}$ to zero\\
        Remove the filters of momentum tensor $\textbf{M}$ using the same index as $R_h$\\
        Set the filters of momentum tensor $\bf{M}$ to zero using the same index as $R_s$\\
    }
    Evaluate the model\\
}

\end{algorithm}

\vspace{-0.3cm}
\section{Experiments}

Our experiments consider three different visual recognition tasks: (1) image classification (2) object detection and (3) domain adaptation. 
For image classification we evaluate our approach on MNIST and CIFAR10 datasets with several commonly used network architectures. Additionally, we conduct an ablation study to better understand which parts of our approach are important for good performance.
For object detection we progressively prune the VGG16 backbone of Faster R-CNN and show a good trade-off between detection performance and computation. 
Finally, for domain adaptation, we evaluate our method on the Office-31 datasets. As our method is progressive, we can prune and adapt to the target domain simultaneously, which is beneficial.
The source code for our paper will be available at \url{https://github.com/Anon6627/Pruning-PGP}.



\subsection{Classification}

In this section, we compare the experimental results obtained using the proposed PGP and RPGP techniques against state-of-the-art filter pruning techniques that are representative of each family described in Section 2.2: L1-norm Pruning (prunes once), Taylor Pruning  (prunes iteratively), DCP (specialised loss function and minimize reconstruction error) and PSFP (progressive pruning). 
Performance is measured in terms of accuracy, and in terms of memory and time complexity (number of parameters and number of FLOPS). For techniques like our PGP, and PSFP, DCP and L1, it is possible to set a target pruning rate $t_\text{prune}$ hyper parameter. For a fixed pruning rate, the complexity (number of FLOPS and parameters) is identical for these techniques, so we can compare them in terms of accuracy for a given complexity. In contrast, techniques like Taylor prunes some filters at each iteration and can be stopped at a certain condition like number of FLOPs or until 99\% of the filters are pruned.
Due to the skip connections, pruning ResNet needs a special strategy. We decided to follow the popular pruning strategy proposed in \cite{L1Pruning} -- pruning the down-sampling layer and then using the same indexes to prune the last convolution of the residual.  Techniques are compared using Resnet, LeNet and VGG networks trained to address benchmark problems.

\paragraph{MNIST:} 
On this dataset, we use the same hyper-parameters as in the original papers. The same settings were used for LeNet5 and ResNet20. With PGP and RPGP, we use a learning rate 0.01, momentum 0.9, 40 epochs with a remove rate of 50\%. For PSFP, we used these same settings except for removal rate of 50\%. 
For Taylor \cite{Molchanov}, we iteratively remove 5 filters each time, and then fine-tune for 5 epochs. This varies slightly from the original procedure because this configuration does not collapse and return the best result. For L1 pruning, we use a 20 epochs fine-tuning after pruning.  For DCP, we ran the author's code for MNIST over 40 epochs, with 20 epochs for the filter pruning and 20 epochs for fine-tuning. 
%
\begin{table}[!t]
\caption{Performance of pruning methods for training LeNet5 on the MNIST classification dataset.}
\label{Lenet5 on Mnist}
\resizebox{\columnwidth}{!}{
\begin{tabular}{|ll||r|r|r|}

\hline
\textbf{Methods}     & \textbf{$t_{\mbox{pruned}}$}  & \textbf{Params} & \textbf{FLOPS}      & \textbf{Error \% ($\pm$ gap)}    \\ \hline \hline
Baseline LeNet5            & \ \ 0\%               & 61K    & 446k       & 0.84 \ \enspace \ \  \enspace \ \  (\  0)      \\ \hline
\hline

\multirow{3}{*}{L1 \cite{L1Pruning}}       & 30\%              & 34.1K  & 304K  & 0.9 \ \ (\ +0.06)  \\ \cline{3-5} 
                          & 50\%              & 18K    & 152K  & 1.05 \ \  (\ +0.21)   \\ \cline{3-5} 
                          & 70\%              & 84K    & 82K   & 2.22 \ \ (\ +1.38)   \\ \cline{3-5}  \hline
\multirow{3}{*}{Taylor \cite{Molchanov}}   & 30\%              & 38K    & 286K  & 0.9  \ \ (\ +0.06)    \\ \cline{3-5} 
                          & 50\%              & 24K    & 76K  & 1.05 \ \ (\ +0.21)   \\ \cline{3-5} 
                          & 70\%              & 13K    & 34K  & 1.22 \ \  (\ +0.38)   \\ \cline{3-5} \hline
\multirow{3}{*}{DCP \cite{Discrimination-aware}}      & 30\%             & 42.7K      & 325K     & 2.75 \ \ (\ +1.91)             \\ \cline{3-5} 
                          & 50\%              & 30.5K      & 232K     & 4.18 \ \ (\ +3.34)            \\ \cline{3-5} 
                          & 70\%\footnotemark              & 30.5K      & 232K     & 6.28 \ \ (\ +5.44)            \\ \cline{3-5}  \hline
\multirow{3}{*}{PSFP \cite{PSP}}     & 30\%              & 34.1K  & 304K  & 1.32 \ \ (\ +0.48)   \\ \cline{3-5} 
                          & 50\%              & 18K    & 152K  & 2.27 \ \ (\ +1.43)   \\ \cline{3-5} 
                          & 70\%              & 84K    & 82K   & 2.99 \ \  (\ +2.15)   \\ \cline{3-5}  \hline \hline
\multirow{3}{*}{PGP\_GN\textsubscript{G} (ours)}  & 30\%     & 34.1K  & 304K  & 0.87 \ \ (\ +0.03)   \\ \cline{3-5} 
                          & 50\%              & 18K    & 152K  & 1.08 \ \  (\ +0.24)   \\ \cline{3-5} 
                          & 70\%              & 84K    & 82K   & 1.74  \ \ \ \ (\ +0.9)    \\ \cline{3-5}  \hline
\multirow{3}{*}{RPGP\_GN\textsubscript{S} (ours)} & 30\%     & 34.1K  & 304K  & 0.9 \ \  (\ +0.06)    \\ \cline{3-5} 
                          & 50\%              & 18K    & 152K  & 1.25 \ \ (\ +0.41)   \\ \cline{3-5} 
                          & 70\%              & 84K    & 82K   & 1.75 \ \  (\ +0.91)   \\ \cline{3-5}  \hline
\end{tabular}
}
\vspace{-4mm}
\end{table}
\footnotetext{Since DCP's code, provided by the authors, did not handle non-residual architecture, we had to modified the original code. Pruning rate above 50\% are struck on LeNet and VGG19}
\begin{table}[!t]
\caption{Performance of pruning methods for training ResNet20 on the MNIST classification dataset.}
\label{Resnet20 on Mnist}
\resizebox{\columnwidth}{!}{
\begin{tabular}{|ll||r|r|r|}
\hline
\textbf{Methods}     & \textbf{$t_{\mbox{pruned}}$}   & \textbf{Params} & \textbf{FLOPS}      & \textbf{Error \% ($\pm$ gap)}    \\ \hline \hline
Baseline Resnet20            & \ \ 0\%               & 272K    & 41M       & 0.74 \ \ \  \enspace \enspace \ \ (\  0)      \\ \hline
\hline
\multirow{3}{*}{L1 \cite{L1Pruning}}       & 30\%              & 137K   & 22M   & 0.75 \ \  (\ +0.01)   \\ \cline{3-5} 
                          & 50\%              & 68K    & 10M   & 1.09 \ \ (\ +0.35)   \\ \cline{3-5} 
                          & 70\%              & 27K    & 4.2M  & 2.02 \ \ (\ +1.28)   \\ \cline{3-5} \hline
\multirow{3}{*}{Taylor \cite{Molchanov}}   & 30\%              & 149K   & 17.7M & 0.87 \ \ (\ +0.13)   \\ \cline{3-5} 
                          & 50\%              & 87K   & 7.8M & 0.95 \ \ (\ +0.21)   \\ \cline{3-5} 
                          & 70\%              & 36K   & 2.6M & 1.04 \ \  (\ +0.30)    \\ \cline{3-5}  \hline
\multirow{3}{*}{DCP \cite{Discrimination-aware}}      & 30\%              & 193K   & 30.3M & 1.11 \ \ \ (+0.37)   \\ \cline{3-5} 
                          & 50\%              & 138K   & 21.1M & 0.62 \ \ (\ -0.12)  \\ \cline{3-5} 
                          & 70\%              & 87.7K  & 13.5M & 1.19 \ \  (\ +0.45)   \\ \cline{3-5} \hline
\multirow{3}{*}{PSFP \cite{PSP}}     & 30\%              & 137K   & 22M   &  0.5 \ \ \  (\ -0.24) \\ \cline{3-5} 
                          & 50\%              & 68K    & 10M   & 0.61 \ \  (\ -0.13)   \\ \cline{3-5} 
                          & 70\%              & 27K    & 4.2M  & 0.72 \ \   (\ -0.02)  \\ \cline{3-5}  \hline \hline
\multirow{3}{*}{PGP\_GN\textsubscript{G} (ours)}  & 30\%              & 137K   & 22M   & 0.4 \ \ \    (\ -0.34)   \\ \cline{3-5} 
                          & 50\%              & 68K    & 10M   & 0.51 \ \  (\ -0.23)  \\ \cline{3-5} 
                          & 70\%              & 27K    & 4.2M  & 0.57 \ \  (\ -0.17)  \\ \cline{3-5}  \hline
\multirow{3}{*}{RPGP\_GN\textsubscript{S} (ours)} & 30\%              & 137K   & 22M   & 0.4 \ \ (\ -0.34)  \\ \cline{3-5} 
                          & 50\%              & 68K    & 10M   & 0.48 \ \ (\ -0.29)  \\ \cline{3-5} 
                          & 70\%              & 27K    & 4.2M  & 0.5 \ \ (\ -0.24)  \\ \cline{3-5} \hline
\end{tabular}
}
\vspace{-4mm}
\end{table}
Results in Tab. \ref{Lenet5 on Mnist} show that our PGP methods compare favorably against State-of the-art   techniques like L1,Taylor and PSFP. Similar tendencies are seen in Tab. \ref{Resnet20 on Mnist}. We also see that PGP performs slightly better than DCP in some case. Finally, since both PGP\_GN\textsubscript{G} and RPGP\_GN\textsubscript{S} have the same criterion, results show that their procedure that differs. The slight better performance of PGP\_GN\textsubscript{G} can be explained by the fact that the pruning criterion is calculated using Batch Gradient Descent instead of Stochastic Gradient Descent.

\vspace{-0.3cm}
\paragraph{CIFAR10:}
In this case, we use a VGG19 for CIFAR10, with learning rate 0.1, momentum 0.9, 400 epochs and we decrease the learning rate by a factor of 10 at 160 and 240 epochs. We also use Resnet56 adapted to CIFAR10 with the same settings, except with 500 epochs. As of PGP and RPGP, we set the remove rate hyper-parameter $r$ to 0.5 (50\%), fine-tune them for 100 epochs after pruned, and store the best score. We use the same settings for PSFP except the removal rate $r$. For Taylor, 5 filters are iteratively iteratively each time and fine-tune on 5 epochs after that. We slightly changed the procedure compared to the original paper because the original procedure pruned one feature map each iteration which is inefficient on a large model. Empirically, we found that 5 feature maps has the best accuracy. For L1 pruning, 100 epochs of fine-tuning are used after pruned to find the best score. With DCP, the settings are provided by the original authors are found to have the best performance.
\begin{table}[!t]
\caption{Performance of pruning methods for training VGG19 on the CIFAR10 classification dataset.}
\label{VGG19 Cifar10}
\resizebox{\columnwidth}{!}{
\begin{tabular}{|ll||r|r|r|}
\hline
\textbf{Methods}     & \textbf{$t_{\mbox{pruned}}$}     & \textbf{Params} & \textbf{FLOPS}      & \textbf{Error \% ($\pm$ gap)}    \\ \hline \hline
Baseline VGG19            &  \ \ 0\%                  & 20M    & 400M       & 6.23 \enspace \enspace \enspace \enspace \ \ \ \  (0)      \\ \hline \hline
\multirow{3}{*}{Li \cite{L1Pruning}} & 30\%     & 9M     & 198M       & 16.94 \ \ ( \ +8.41)  \\ \cline{3-5} 
                          & 50\%                & 5M     & 100M       & 16.51 \ \ ( \ +7.98)  \\ \cline{3-5} 
                          & 70\%                & 1M     & 37M        & 16.17 \ \ ( \ +7.64)  \\ \cline{3-5}  \hline
\multirow{3}{*}{Taylor \cite{Molchanov}}& 30\%  & 10M   & 156M     & 9.82  \ \ \ ( \ +2.29)  \\ \cline{3-5} 
                          & 50\%               & 5M     & 72M         & 11.94 \ \ ( \ +3.41) \\ \cline{3-5} 
                          & 70\%               & 1.9M    & 24M        & 16.85  \ \ ( \ +8.32)   \\ \cline{3-5} \hline
\multirow{3}{*}{DCP \cite{Discrimination-aware}}  & 30\%              & 10M      & 221M          & 5.8 \ \ ( \ -0.65)             \\ \cline{3-5} 
                          & 50\%              & 6M      & 158M & 7.76 \ \ ( \ +1.53)     \\ \cline{3-5} 
                          & 70\%\footnotemark[\value{footnote}]             & 6M      & 158M          &  7.86 \ \ ( \ +1.63)            \\ \cline{3-5}  \hline
\multirow{3}{*}{PSFP \cite{PSP}}        & 30\%    & 9M     & 198M       & \ 8.98 \ \  ( \ +2.75)  \\ \cline{3-5} 
                          & 50\%              & 5M     & 100M           & 11.2  \ \  ( \ +4.97)  \\ \cline{3-5} 
                          & 70\%              & 1M     & 37M            & 12.06 \ \ ( \ +5.83)  \\ \cline{3-5} \hline \hline

\multirow{3}{*}{PGP\_GN\textsubscript{G} (ours)}  & 30\%        & 9M     & 198M             & \ 7.37 \ \  ( \ +1.14)   \\ \cline{3-5} 
                          & 50\%              & 5M     & 100M           & \ 8.38  \ \ ( \ +2.15)  \\ \cline{3-5} 
                          & 70\%             & 1M     & 37M             & \ 9.7 \ \ ( \ +3.47)  \\ \cline{3-5}  \hline
\multirow{3}{*}{RPGP\_GN\textsubscript{S} (ours)} & 30\%         & 9M     & 198M            & \ 7.65 \ \ ( \ +1.42)   \\ \cline{3-5} 
                          & 50\%              & 5M     & 100M           & \ 8.79\ \ ( \ +2.56)  \\ \cline{3-5} 
                          & 70\%              & 1M     & 37M            & 10.56  \ \ ( \ +4.33)  \\ \cline{3-5}  \hline
\end{tabular}
}
\vspace{-4mm}
\end{table}
\begin{table}[]
\caption{Performance of pruning methods for training ResNet56 on the CIFAR10 classification dataset.}
\label{Resnet56 on Cifar10}
\resizebox{\columnwidth}{!}{
\begin{tabular}{|ll||r|r|r|}
\hline
\textbf{Methods}     & \textbf{$t_{\mbox{pruned}}$}   & \textbf{Params} & \textbf{FLOPS}      & \textbf{Error \% ($\pm$ gap)}    \\ \hline \hline
Baseline Resnet56            & \ \ 0\%               & 855K    & 128M       & \ 6.02 \enspace \enspace \enspace   \ \ \ \ ( 0)      \\ \hline \hline
\multirow{4}{*}{L1 \cite{L1Pruning}}       & 30\%              & 431K   & 67M   & 13.34 \ \ (\ +7.32)      \\ \cline{3-5} 
                                          & 50\%              & 215K   & 32M   & 15.57 \ \ (\ +9.55)      \\ \cline{3-5} 
                                          & 70\%              & 84K    & 13M    & 17.89  (\ +11.87)      \\ \cline{3-5}  \hline
\multirow{4}{*}{Taylor \cite{Molchanov}}   & 40\%              & 491K    & 51M & 13.9 \ \ \ \  (\ +7.88)      \\ \cline{3-5} 
                                      & 50\%              & 268K    & 23M & 15.34 \ \ (\ +9.32)      \\ \cline{3-5} 
                                      & 70\%             & 100k    & 8M & 22.1 \ \ (\ +16.08)      \\ \cline{3-5}  \hline
\multirow{4}{*}{DCP \cite{Discrimination-aware}}      & 30\%               & 600K     & 90M      & 5.67 \ \ \ \ (\ -0.35)          \\ \cline{3-5} 
                          & 50\%              & 430K      & 65M      & 6.43 \ \  \ (\ +0.41)          \\ \cline{3-5} 
                          & 70\%              & 270K      & 41M      & 7.18 \ \ \ (\ +1.16)          \\ \cline{3-5}  \hline
\multirow{4}{*}{PSFP \cite{PSP}}     & 30\%              & 431K   & 67M   & 8.94 \ \ \ (\ +2.92)      \\ \cline{3-5} 
                          & 50\%              & 215K   & 32M   & 10.93  \ \  (\ +4.91)      \\ \cline{3-5} 
                          & 70\%              & 84K    & 13M    & 14.18 \ \ (\ +8.16)      \\ \cline{3-5} \hline \hline

\multirow{4}{*}{PGP\_GN\textsubscript{G} (ours)}  & 30\%              & 431K   & 67M   & 8.95 \  \   (\ +2.93)      \\ \cline{3-5} 
                          & 50\%              & 215K   & 32M   & 10.59 \ \ (\ +4.57)      \\ \cline{3-5} 
                          & 70\%              & 84K    & 13M    & 13.02  \ \enspace \enspace  \ \ (\ +7)      \\ \cline{3-5} \hline
\multirow{4}{*}{RPGP\_GN\textsubscript{S} (ours))} & 30\%              & 431K   & 67M   & 9.37  \ \enspace  (\ +3.35)      \\ \cline{3-5} 
                          & 50\%              & 215K   & 32M   & 10.46 \ \ (\ +4.44)      \\ \cline{3-5} 
                          & 70\%              & 84K    & 13M    & 14.16 \ \ (\ +8.14)      \\ \cline{3-5}  \hline
\end{tabular}
}
\vspace{-4mm}
\end{table}
From Tabs. \ref{VGG19 Cifar10} and \ref{Resnet56 on Cifar10}, our techniques consistently perform better than state of the art techniques L1, Taylor and PSFP on VGGNet. For ResNet, PSFP has a different pruning strategy on ResNet, and does not prune the down-sample layer, and therefore does not prune the last convolutional layer of the residual. This translates into a slight better accuracy on some settings. Our ablation study also provides a comparison of techniques using the same pruning strategy on ResNet, and shows the importance of momentum pruning. DCP performs better than ours on this dataset, mainly because of the additional losses that help selecting discriminate filters. However, it is difficult to compare directly since they do not yield  the same number of FLOPS and parameters, and DCP starts from a trained model and requires more computation power.


\subsection{Ablation study}

The \textbf{training and pruning time} of a model are important factors of a technique, for instance for deploying or adapting a model in an operational environment. One of advantage of progressive pruning techniques is the reduction of processing time at each epoch since filters are removed while training, at each epoch. Tab. \ref{Comparison of pruning time 0.9} presents the training and pruning time pruning for the evaluated techniques. For progressive pruning technique, values represent both pruning and training times,  while for DCP, L1 and Iterative pruning, values represent  (training time) + pruning and retrain times. Experiments are conducted on the CIFAR10 dataset with the same settings as above. 
\begin{table}[]
\caption{Training and pruning time for pruning techniques with $t_\text{prune} = 0.5$ and $0.9$.}
\label{Comparison of pruning time 0.9}
\resizebox{\columnwidth}{!}{
\begin{tabular}{lrrrr}
 & \multicolumn{1}{l}{} & \multicolumn{1}{l}{} & \multicolumn{1}{l}{} & \multicolumn{1}{l}{} \\ \hline
\multicolumn{1}{|l||}{\textbf{Methods}} & \multicolumn{2}{c|}{\textbf{VGG19}} & \multicolumn{2}{c|}{\textbf{Resnet56}} \\ 
\multicolumn{1}{|l||}{\ \ $t_{\mbox{pruned}}$} & \multicolumn{1}{c|}{0.5} & \multicolumn{1}{c|}{0.9} & \multicolumn{1}{c|}{0.5} & \multicolumn{1}{c|}{0.9} \\ \hline \hline
\multicolumn{1}{|l||}{Baseline} & \multicolumn{1}{c|}{219m} & \multicolumn{1}{c|}{219m} & \multicolumn{1}{c|}{307m} & \multicolumn{1}{c|}{307m} \\ \hline
\multicolumn{1}{|l||}{L1 \cite{L1Pruning}} & \multicolumn{1}{c|}{(219) \ \ + 32m} & \multicolumn{1}{c|}{(219) \ \ + 32m} & \multicolumn{1}{c|}{(307) \ \  + 48m} & \multicolumn{1}{c|}{(307) \ \  + 48m} \\ \hline
\multicolumn{1}{|l||}{Taylor \cite{Molchanov}} & \multicolumn{1}{c|}{(219) \ \  + 254m} & \multicolumn{1}{c|}{(219) \ \ + 457m} & \multicolumn{1}{c|}{(307) \ \ + 488m} & \multicolumn{1}{c|}{(307) \ \ + 878m} \\ \hline
\multicolumn{1}{|l||}{DCP \cite{Discrimination-aware}} & \multicolumn{1}{c|}{-} & \multicolumn{1}{c|}{-} & \multicolumn{1}{c|}{(307) + 489m} & \multicolumn{1}{c|}{(307) + 443m} \\ \hline
\multicolumn{1}{|l||}{PSFP \cite{PSP}} & \multicolumn{1}{c|}{219m} & \multicolumn{1}{c|}{219m} & \multicolumn{1}{c|}{307m} & \multicolumn{1}{c|}{307m} \\ \hline \hline
\multicolumn{1}{|l||}{PGP (ours)} & \multicolumn{1}{c|}{329m} & \multicolumn{1}{c|}{329m} & \multicolumn{1}{c|}{441m} & \multicolumn{1}{c|}{441m} \\ \hline
\multicolumn{1}{|l||}{RPGP (ours)} & \multicolumn{1}{c|}{\bf{211m}} & \multicolumn{1}{c|}{\bf{168m}} & \multicolumn{1}{c|}{\bf{263m}} & \multicolumn{1}{c|}{\bf{241m}} \\ \hline
\end{tabular}
}
\vspace{-4mm}
\end{table}

From Tab.~\ref{Comparison of pruning time 0.9}, the fastest pruning method (without considering training time) is currently the L1. However, it should be noted that the original training of the model takes around 219 mins for VGG and 307 mins for Resnet56. So, taking into account also training time L1 is slower than our approach. Other techniques likes Taylor prune in a iterative way composed of multiple feature maps and fine-tuning, this method can be very slow, depending on the number of filters pruned at each iteration. DCP is particulary slow since it needs to start from an already trained model and then the pruning process need to do the filter pruning optimization process and the fine-tuning after pruning. For PSFP, this algorithm has similar time to the original training since it does not technically change the size of the model during training. Between PGP and RPGP, the difference is the use of an entire epoch to compute the pruning criterion with PGP, and the direct computation of the criterion  during a training epoch with RPGP. Also, since we hard-prune filters at each epoch, the epoch time will become faster as the model is pruned/trained. Overall, the progressive pruning methods train and prune in considerably less time than other methods. 

To compare the \textbf{selection criterion}, we use the same configuration as the general comparison for RPGP on CIFAR10, except we vary the criterion and set a pruning rate of 50\%. 
\begin{table}[t!]
\centering
\caption{Error rate for RPGP with different pruning criteria.}
\label{Ablation study on criteria}
\resizebox{0.9\columnwidth}{!}{
\small
\begin{tabular}{|l|l|l|l|l|l|}
\hline
\textbf{Networks}   & \textbf{L2}   & \textbf{Taylor} & \textbf{TW} & \textbf{GN\_G} & \textbf{GN\_S} \\ \hline
\hline
VGG19  & \textbf{\ 8.47\%}  & \ 9.27\%            & \ 8.78\%        & \textbf{\ 8.47\%}  & \ 8.79\%           \\ \hline
ResNet56            & 10.30\%          & 10.97\%           & 10.46\%       & \textbf{10.24\%} & 10.28\%          \\ \hline
\end{tabular}
}
\vspace{-4mm}
\end{table}
In Tab.~\ref{Ablation study on criteria}, we can see that our criterion performs better than others in the context of progressive pruning, and similar to the L2 Norm. The comparison between Taylor Weight (TW), and Gradient Norm (GN) shows that a small gradient norm during training may be a good indicator about the importance of a filter. From the table we can also see that Taylor Weights performs better than the original Taylor criterion. Overall $GN_G$, which uses batch gradient to capture changes, seems to work the best with progressive pruning. As for the similarity between L2 and $GN$, it is explained in the Supplemental Material.

In this experiment of momentum pruning, the same strategy, hyper-parameters and L2 criterion are used for both RPGP and PSFP. The only difference is that RPGP performs \textbf{momentum pruning}.
\begin{table}[t!]
\centering
\caption{Error rates for RPGP and PSFP with L2.}
\label{Ablation study on pruning procedure}
\resizebox{0.5\columnwidth}{!}{
\begin{tabular}{|l|l|l|}
\hline
\textbf{Method} & \textbf{VGG19} & \textbf{ResNet56} \\ \hline \hline
PSFP    & 11.20\%           & 10.93\%                   \\ \hline
RPGP    & \textbf{\ 8.47\%}   & \textbf{10.09\%}         \\ \hline
\end{tabular}
}
\vspace{-4mm}
\end{table}
From the Tab. \ref{Ablation study on pruning procedure}, in both of the case (VGG19 and ResNet56), our proposed methods performs better than the state of the art PSFP method. Since, everything is the same in this setting except the momentum pruning, this clearly shows the advantage of pruning momentum during progressive pruning.
%




As described, PSFP does not \textbf{prune the downsampling layer} of ResNet56, thus, it does not prune the last layer of the residual connection. The performance of PSFP and RPGP is compared using the same strategy on ResNet56, i.e., the downsampling layer and last layer of residual connection are not pruned, on with CIFAR10 dataset and the same hyper-parameters as in previous experiments.
The results in Tab. \ref{Ablation study on pruning resnet procedure} indicate that the RPGP approach typically performs better than PSFP. Interestingly, when no pruning is performed on the downsampling layer and last layer of the residual connection, our method performs much better. The residual connection is sensitive to pruning, and may require a different pruning strategy.
\begin{table}[h]
\vspace{-0.4cm}
\caption{Error rates of PSFP and RPGP with different pruning rates, when downsampling and last layers of residual connection are not pruned.}
\label{Ablation study on pruning resnet procedure}
\centering
\resizebox{0.7\columnwidth}{!}{
\begin{tabular}{|l|l|l|l|l|}

\hline
        & \multicolumn{4}{c|}{ \textbf{$t_\text{prune} \times 100$\%}} \\ 
\textbf{Methods} & 30\%  & 50\%  & 70\%   & 90\%   \\ \hline \hline
PSFP    & 8.94  & 10.93   & 14.18  & 28.09  \\ \hline
RPGP(GN\_S)    & \textbf{8.87}  & \textbf{10.09}   & \textbf{11.02}  & \textbf{13.94}  \\ \hline
\end{tabular}
}
\vspace{-4mm}
\end{table}

\subsection{Detection}
\paragraph{PASCAL VOC:} In this case, PGP, RPGP and PSFP  techniques are adapted for an object detection problem. We progressively prune a Faster R-CNN with a VGG16 backbone using a learning rate of 0.001, momentum of 0.9 using a 10 epochs progressive pruning, and early stopping for fine-tuning over a few epochs. For the L1 pruning, a trained model is prune 50\% from the network, and then we fine-tune on Pascal VOC.  For this experiment, we set the pruning rate hyper-parameter $r$ to 0.5 (50\%), and show mean average precision (MAP) measure for comparison. 
In this experiment we skip the pruning of the last layer since it would mean pruning the input of the RPN layer, which we empirically found that it results in significant performance reduction.
In Tab.\ref{Comparison of detection}, PGP and RPGP perform better than PSFP, the current state-of-the-art progressive pruning. However, the PGP needs more time to prune due to the calculation of the criterion in a separate epoch. RPGP provides a slightly better performance (possibly due to stochasticity), and with much less pruning time. The difference in accuracy between RPGP and PSFP highlights the importance of momentum pruning with these approaches. The significant difference in the training time between RPGP and PSFP also suggests that by adding hard pruning to existing soft pruning during training can reduce training time. 

\begin{table}[]
\vspace{-7mm}
\caption{Performance of pruning methods for training Faster R-CNN with VGG16 backbone on the Pascal VOC detection dataset with $t_\text{pruned}=50\%$.}
\label{Comparison of detection}
\centering
\resizebox{0.9\columnwidth}{!}{

\begin{tabular}{|l|r|r|l|l|}
\hline
\textbf{Methods}            & Params & FLOPS & mAP    & Training Time \\ \hline \hline
Baseline VGG16     & 137M   & 250G  & 69.6\% & 428m                     \\ \hline
L1  \cite{L1Pruning}               & 125M   & 174G  & 62.3\% & (428) + 31m   \\ \hline
PSFP \cite{PSP} & 125M   & 174G  & 63.5\% & 428m                        \\ \hline \hline
PGP\_GN\textsubscript{G} (ours)  & 125M   & 174G  & 65.5\% & 769m       \\ \hline
RPGP\_GN\textsubscript{S} (ours) & 125M   & 174G  & \textbf{66.0}\% & \textbf{281m}          \\ \hline
\end{tabular}
}
\vspace{-4mm}
\end{table}

\subsection{Domain Adaptation}

\paragraph{Office-31:} We adapted our pruning technique for domain adaptation. Among unsupervised domain adaptations techniques\cite{GRL, DAN, ADDA}, we chose Deep Adaptation Network(DAN)\cite{DAN} with a VGG16 backbone since it is a popular technique. 
However, we could adapt our pruning technique to other unsupervised domain adaptation such as ADDA\cite{ADDA} or DANN\cite{GRL}. For this experiment, we train our model for 400 epochs for joint domain-adaptation and pruning, with an exponential decay learning rate starting at 0.001, we set a pruning rate of 20\% with a remove rate of 0.3.
In Tab.\ref{Comparison of Domain Adaptation} we see that our method performs better than TCP in the majority of the cases while having a higher FLOPs Reduction. We can explain the improvement over TCP mainly due to two factors. First, the use of our pruning criterion. As shown in our ablation study, a Taylor criterion based on weight is better than its feature map counterpart. Second, we think that progressive pruning is more suited for domain adaptation than the commonly used scheme in which first the model is adapted to the target domain, then pruned, and finally fine-tuned. In our approach we perform pruning and domain adaptation simultaneously in a single training. In this way, instead of trying to adapt all filters of the model, we might expect to directly remove those filters that are specific of a certain domain and do not help in bridging domain discrepancy. This would ease the learning of transferable features in the case of DAN\cite{DAN} or domain confusion in the case of ADDA\cite{ADDA}.

\begin{table}[]
\caption{Performance of our methods for training DAN with VGG16 backbone on Office-31 with $t_\text{pruned}=20\%$.}
\label{Comparison of Domain Adaptation}
\resizebox{\columnwidth}{!}{
\begin{tabular}{|l|l|l|l|l|}
\hline
                   & Source-only & Baseline VGG & TCP   & RPGP\_GN\textsubscript{S}(ours)        \\ \hline
FLOPS Reduction \% & 0\%         & 0\%          & 26\%  & \textbf{35\%}        \\ \hline
$A\rightarrow W$           & 68.5        & 74.0           & 76.1  & \textbf{78.2}        \\ \hline
$A\rightarrow D$           & 61.1        & 72.3         & 76.2  & \textbf{77.7}        \\ \hline
$W\rightarrow A$           & 41.6        & 55.2         & 51.2  & \textbf{51.6}        \\ \hline
$W\rightarrow D$           & 94.3        & 97.5         & \textbf{99.8}  & 99.4        \\ \hline
$D\rightarrow W$           & 94.5        & 94.0           & 96.1  & \textbf{96.5}        \\ \hline
$D\rightarrow A$           & 50.3        & 54.1         & 47.9  & \textbf{48.0}          \\ \hline
Average            & 68.3 & 74.5  & 74.5 & \textbf{75.2} \\ \hline
\end{tabular}
}
\vspace{-4mm}
\end{table}

%

\vspace{-0.25cm}
\section{Conclusion}

In this paper, we show that it is possible to efficiently prune a deep learning model from scratch with the PGP technique while improving the trade-off between compression, accuracy and training time. PGP is a new progressive pruning technique that relies on change in filter weights to apply hard and soft pruning strategies that allows for pruning along the back-propagation path. The filter selection criterion is well adapted for progressive pruning from scratch when the norm of the gradient is considered.
Results obtained from pruning various CNNs on image data for classification and object detection problems show that the proposed PGP allows maintaining a high level of accuracy with compact networks. Results show that PGP can achieve better CNN optimisations than PSFP, often translating to a higher level of accuracy for a same pruning rate as PSFP and other state-of-art techniques. In domain adaptation problems, our technique outperforms current state of the art technique while pruning more. Future research will involve analyzing the performance of different CNNs pruned using the proposed method on larger datasets from real-world visual recognition problems (e.g., tracking and recognition of persons in video surveillance) and on different domain adaptation architetures.

{\small
\bibliographystyle{ieee}
\bibliography{egbib}
}

\newpage
\appendix

\clearpage
{ \centering \Large \textbf{Supplementary Material}}

\section{Additional Experimental Results}

\subsection{Implementation Details}

One of the problem of pruning during training is how to handle the shape of the gradient tensor and momentum tensor during backward pass. In the case of PyTorch \cite{PyTorch}, the shape of the gradient tensor and momentum tensor is usually handled by the optimizer, which does not necessary update the shape during forward pass. Also, redefining a new optimizer with the new pruned model in a trivial way would result in losing all values accumulated in the momentum buffer. One of the way to overcome this, is to prune also the gradient and momentum tensors using indexes that we used to prune the weight tensor, and then transfer them to a newly defined optimizer. 

\subsection{Graphical comparison on CIFAR10 with VGG:}

The results presented in this section are similar to the ones shown in Tabs. \ref{Lenet5 on Mnist} to \ref{Resnet56 on Cifar10} of our paper. In the main paper, we could only compare the performance of methods with 4 pruning rates due to space constraints. In this section, we compare the performance of methods using the same experimental settings (as in our paper), but with 10 data points ($t_{\mbox{pruned}} = 0.1, 0.2, ..., 1.0$) on L1 \cite{L1Pruning}, Taylor \cite{Molchanov}, PSFP \cite{PSP} and our approach. 
Since the number of remaining parameters can differ slightly from one algorithm to the other, some of the value on X-axis are rounded up for a better visualization.

\begin{figure}[h!]
    \centering
    \includegraphics[width=9cm]{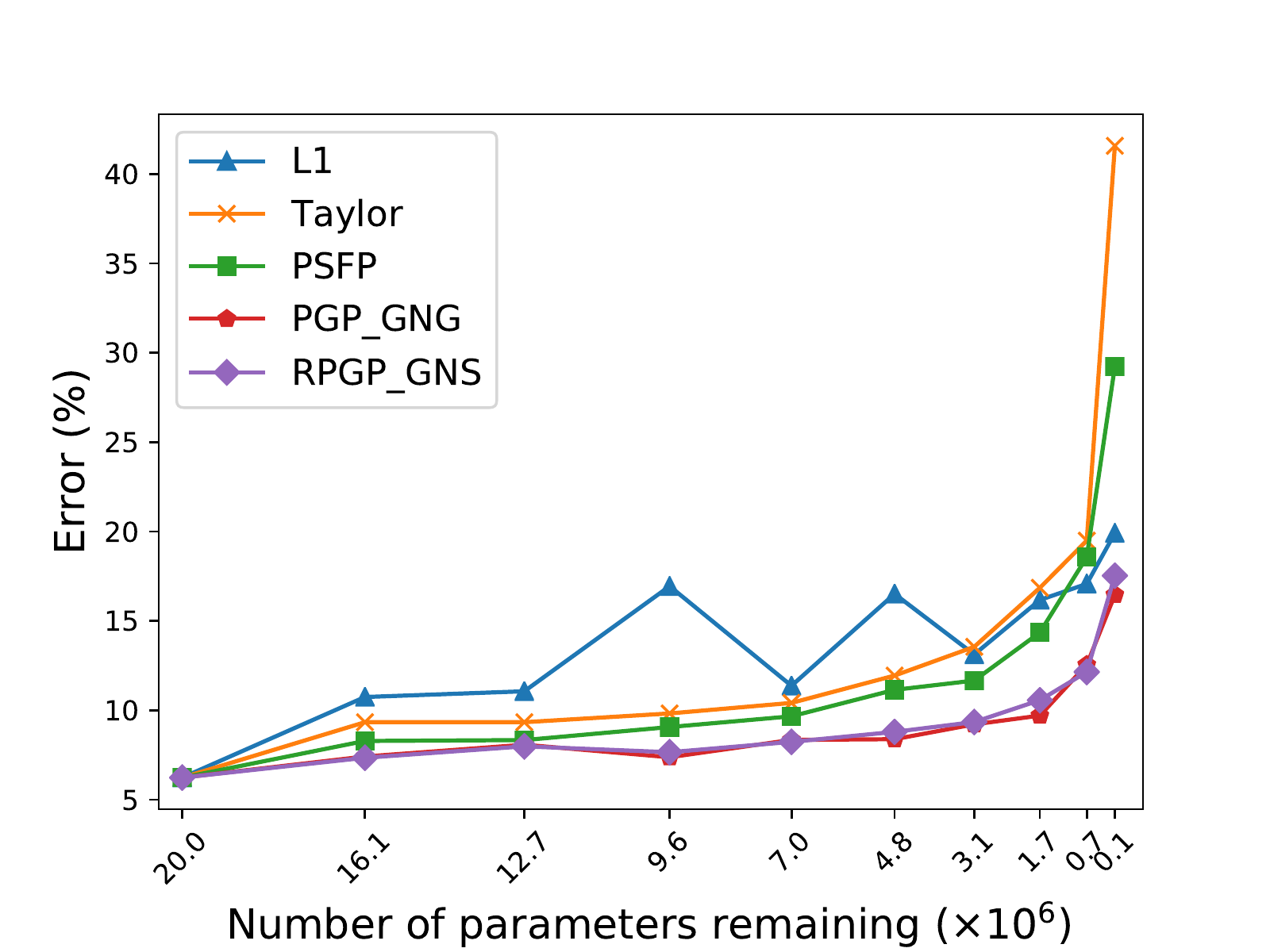}
    \caption{Error rate versus the number of remaining parameters with the proposed and baseline pruning methods for VGG19 on the CIFAR10 dataset.}
    \label{More comparison}
    \vspace{-4mm}
\end{figure}

Results in Figure \ref{More comparison} show the proposed PGP and RPGP pruning methods consistently outperforming the other methods. Note that the proposed methods allow to  maintain a low lever of error event with an important increase in the pruning rate.

\subsection{L2 vs Gradient Norm:}

From the ablation study, we noticed that the performance of L2 and Gradient norm is very similar in the case of soft pruning. This can be understood considering the following:
 \begin{equation}\label{L2vsGN}
	\begin{aligned}
	||\textbf{W}^j_{i}||_2 &= \scalebox{1}{$||\textbf{W}_{i}^{j-1} - \alpha \frac{\partial \mathcal{L}^{j-1}}{\partial \textbf{W}_i^{j-1}}||_2$} \\
	&= \scalebox{1}{$||\textbf{W}_{i}^{j-2} - \alpha \frac{\partial \mathcal{L}^{j-2}}{\partial \textbf{W}_i^{j-2}} - \alpha \frac{\partial \mathcal{L}^{j-1}}{\partial \textbf{W}_i^{j-1}}||_2$} \\
	&= \scalebox{1}{$||\textbf{W}_{i}^{0} - \alpha \sum_{k=0}^{j}\frac{\partial \mathcal{L}^{k}}{\partial \textbf{W}_i^{k}}||_2$}
	\end{aligned}
\end{equation}
Where $\textbf{W}^j_i$ represents the weight of an filter $i$ at iteration $j$ in an epoch, $\alpha$ is the learning rate, and $\mathcal{L}^k$ denotes here the loss function at iteration $k$. From the Equ.\ref{L2vsGN} we can observe the difference between L2 and Gradient Norm 
is the initial values of $\textbf{W}^0_i$. Taking in account the partial soft pruning nature of our approach, $\textbf{W}^0_i$ can be zero when it is soft pruned.  Therefore the two approaches tends to have similar values (since $\alpha$ is a scalar, it is not important in this context).

\subsection{Progressive pruning from scratch vs trained:}

Tab.~\ref{pretraining} shows that the performance obtained by a model that was randomly initialized (scratch) versus one that was pre-trained on CIFAR10 using the same settings as before ($t_{pruned} = 50\%$, $r =0.5$).
\begin{table}[t!]
\caption{Error rate for RPGP when trained from scratch compared to a trained model.}
\label{Ablation study RPGP(from scratch) and RPGP(from trained model)}
\centering
\resizebox{0.8\columnwidth}{!}{
\begin{tabular}{|l||c|c|}
\hline
\textbf{Training Scenario}       & \textbf{VGG19}    & \textbf{ResNet56}         \\ \hline \hline
Scratch             & 8.79 \%       & 10.46 \%               \\ \hline
Pre-trained         & \textbf{8.23} \%      & \textbf{\ 9.51} \%               \\ \hline
\end{tabular}
}
\label{pretraining}
\vspace{-4mm}
\end{table}
From Tab. \ref{Ablation study RPGP(from scratch) and RPGP(from trained model)} the difference in terms of accuracy between a network pruned starting from scratch and a network pruned after training is quite reduced and can vary depending on the architectures. Overall, instead of starting from a trained model and prune, the proposed techniques can attain similar performance starting from a randomly initialized model, thus, with a reduced training and pruning time, therefore more suitable for fast deployment. 

\subsection{Hard vs soft pruning:}
RPGP is used with our gradient criterion and a target prune rate at 50\% and using the same hyper-parameters. The removal rate $r$ is varied in order to see the impact of having more or less recovery.
\begin{table}[t!]
\caption{Error rate for RPGP for different removal rates $r$.}
\label{Ablation study on removal rate}
\centering
\resizebox{0.8\columnwidth}{!}{

\begin{tabular}{|l||c|c|c|c|}
\hline
\textbf{Networks}    & $r=0.3$           & $r=0.5$           & $r=0.7$       & $r=1.0$   \\ \hline \hline
VGG19               & \textbf{8.74}\%   & 8.79\%            & 8.99\%        & 8.92\%    \\ \hline
ResNet56            & 10.57\%           & \textbf{10.46}\%  & 11.03\%       & 10.78\%   \\ \hline
\end{tabular}
}
\vspace{-4mm}
\end{table}
The results in Tab. \ref{Ablation study on removal rate} show that a remove rate of 0.3(30\%)or 0.5(50\%) has the best balance between the amount of hard pruning soft pruning. It is also interesting to see that, without any soft pruning ($r$=1.0), the performance of the approach is still close to others removal rate.

\end{document}